\documentclass[sigconf]{acmart}
\renewcommand\footnotetextcopyrightpermission[1]{}
\AtBeginDocument{%
  \providecommand\BibTeX{{%
    \normalfont B\kern-0.5em{\scshape i\kern-0.25em b}\kern-0.8em\TeX}}}

\acmConference[Arxiv]{..}{cs.CV}{2022}
\usepackage{graphicx}
\usepackage{multirow}
\graphicspath{
    {./fig/}
    }

\begin{document}

\title{Dynamic Neural Textures: Generating Talking-Face Videos with Continuously Controllable Expressions}


\author{Zipeng Ye}
\email{yezp17@mails.tsinghua.edu.cn}
%\orcid{0000-0002-4322-7550}
\affiliation{%
  \institution{Tsinghua University}
  \city{Beijing}
  \country{China}
  \postcode{43017-6221}
}

\author{Zhiyao Sun}
\email{sunzy21@mails.tsinghua.edu.cn}
\affiliation{%
  \institution{Tsinghua University}
  \city{Beijing}
  \country{China}
  \postcode{43017-6221}
}

\author{Yu-Hui Wen}
\email{wenyh1616@tsinghua.edu.cn}
\affiliation{%
  \institution{Tsinghua University}
  \city{Beijing}
  \country{China}
  \postcode{43017-6221}
}

\author{Yanan Sun}
\email{sunyn20@mails.tsinghua.edu.cn}
\affiliation{%
  \institution{Tsinghua University}
  \city{Beijing}
  \country{China}
  \postcode{43017-6221}
}

\author{Tian Lv}
\email{lvt18@mails.tsinghua.edu.cn}
\affiliation{%
  \institution{Tsinghua University}
  \city{Beijing}
  \country{China}
  \postcode{43017-6221}
}

\author{Ran Yi}
\email{ranyi@sjtu.edu.cn}
\affiliation{%
  \institution{Shanghai Jiao Tong University}
  \city{Shanghai}
  \country{China}
  \postcode{43017-6221}
}

\author{Yong-Jin Liu}
\email{liuyongjin@tsinghua.edu.cn}
\affiliation{%
  \institution{Tsinghua University}
  \city{Beijing}
  \country{China}
  \postcode{43017-6221}
}

%\renewcommand{\shortauthors}{Anonymous Submission}

\begin{abstract}
Recently, talking-face video generation has received considerable attention. So far most methods generate results with neutral expressions or expressions that are implicitly determined by neural networks in an uncontrollable way. In this paper, we propose a method to generate talking-face videos with continuously controllable expressions in real-time. Our method is based on an important observation: In contrast to facial geometry of moderate resolution, most expression information lies in textures. Then we make use of neural textures to generate high-quality talking face videos and design a novel neural network that can generate neural textures for image frames (which we called {\it dynamic neural textures}) based on the input expression and continuous intensity expression coding (CIEC). Our method uses 3DMM as a 3D model to sample the dynamic neural texture. The 3DMM does not cover the teeth area, so we propose a teeth submodule to complete the details in teeth. Results and an ablation study show the effectiveness of our method in generating high-quality talking-face videos with continuously controllable expressions. We also set up four baseline methods by combining existing representative methods and compare them with our method. Experimental results including a user study show that our method has the best performance.
\end{abstract}





\maketitle

\section{Introduction}
\begin{figure*}
\centering
  \includegraphics[width=\textwidth]{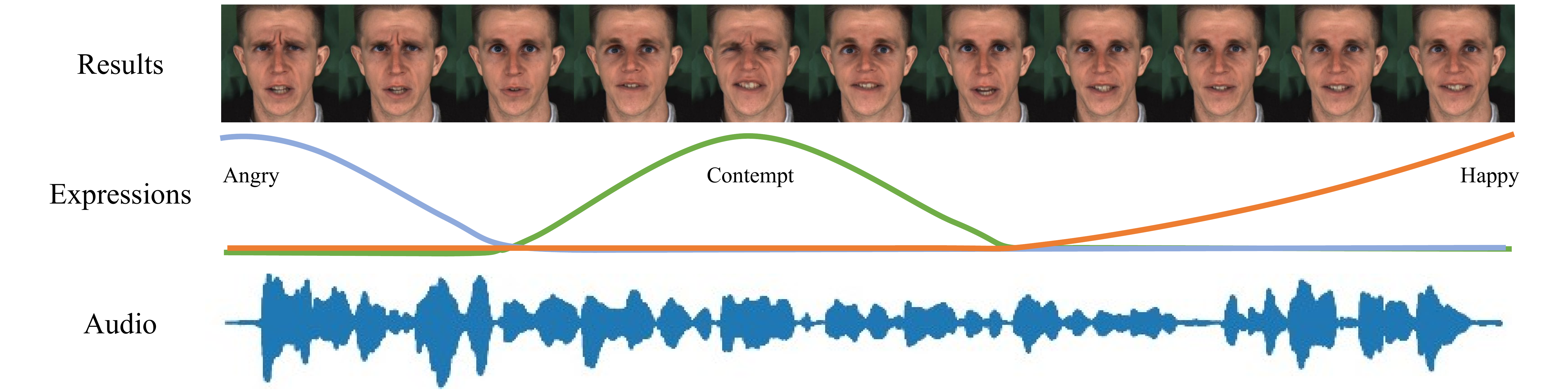}
  \caption{We propose dynamic neural textures to generate talking-face videos with continuously controllable expressions. The input of continuous expression information (class label and intensity) can be specified by user or automatically extracted from input audio~\cite{tao2018ensemble}. Our method switches different expressions at zero intensity, i.e., neutral expression. }
  \label{fig:teaser}
\end{figure*}

Recently, face image editing with controllable characteristics, such as identity, expression, age, etc, has attracted considerable attention (e.g., \cite{karras2019style,richardson2021encoding, shen2020interfacegan}). Editing these characteristics in videos is even more challenging due to the constraint of inter-frame continuity, and very few works exist. For example, a facial expression editing method is proposed in \cite{ma2019real} which transforms the expressions in talking-face videos into two types: happy and sad. In this method, a lip shape correction and a smoothing post-processing are required to retain lip synchronization with the source audio and temporal smoothness. In this paper, we address the challenging problem of generating talking-face videos with {\it continuously controllable expressions}, given an audio and a sequence of expressions (in different classes)
and intensities; see Fig.~\ref{fig:teaser} for an illustration. Different from \cite{ma2019real}, we consider talking-face video generation with diverse styles of expressions whose intensities are continuously controllable, while simultaneously maintaining lip synchronization. 

Many methods have been proposed for synthesizing talking-face videos (e.g., \cite{thies2020neural, wen2020photorealistic, Wav2Lip}). For example, a class of audio-driven talking-face video generation methods~\cite{wen2020photorealistic, thies2020neural, yi2020audio, wu2021imitating, ji2021audio} connect visual and auditory modalities by using intermediate 3D face models to achieve 3D consistency and temporal stability. So far most methods output talking-face videos with expressions whose types and intensities cannot be explicitly controlled by users~\cite{wen2020photorealistic, thies2020neural, yi2020audio, wu2021imitating}. An exceptional work is \cite{ji2021audio} which can generate talking-face videos with controllable expressions by controlling the intermediate 3D morphable model~(3DMM)~\cite{BlanzV99}. Given an input audio, all the above mentioned methods learn to predict the expression parameters to drive the 3D face animation and then use neural networks to generate photo-realistic videos, where the expression parameters are related to lip motions and expressions of the 3D face. Thus, unsynchronized lip motions as well as inaccurate expressions exist in most of the existing methods.

In this paper, we propose a method called {\it dynamic neural textures}, for the purpose of generating talking-face videos with continuously controllable expressions and lip synchronization. Here, continuously controllable expressions mean that the intensity level of different types of expressions can be continuously changed. Our method is based on an important observation: given a 3D face geometry of moderate resolution\footnote{Although 3D geometry of very high resolution can contain more expression information, it is costly to obtain.}~(e.g., 3DMM), we find that textures but not geometry carry sufficient information for describing expressions. Meanwhile, low-frequency vertex colors cannot provide enough expression information. Therefore we design to use texture maps which contain fine details in high-resolution images to control expressions.

Texture images can be obtained from photo-realistic rendering techniques, including traditional physically based rendering and recently proposed neural rendering \cite{tewari2020state}. However, the traditional rendering pipeline requires high-quality 3D models. Our work considers neural rendering because it is able to synthesize photo-realistic images by using a geometric representation of moderate resolution. %Therefore, the method is time-efficient and can achieve real-time. 
In particular, we pay attention to neural textures \cite{thies2019deferred}, which are learnable high-dimensional feature maps, instead of simple RGB values, to encode high-level appearance information. However, neural textures use fixed textures which cannot represent different expressions and usually lead to unnatural results~\cite{olszewski2017realistic,nagano2018pagan}. In this paper, we propose a dynamic neural texture method, which generates different textures for different expressions.  

Our proposed dynamic neural textures are (1) inferred from expression input and (2) independent from geometry, so that we can decouple lip motion (represented in geometry) from expressions (represented in textures). %In more detail, we infer (1) dynamic neural textures from input expressions and (2) 3D face from input audio and input face frames. Therefore, 
The significant difference from static neural textures is that dynamic neural textures depend on input expressions. Thus, dynamic neural textures are expressive to synthesize expressions with continuous intensities. %, which may be unseen during training. 
%We also infer 3D face from input audio and face frames, and use it to sample the dynamic neural textures from texture space into screen space, followed by a neural network to generate photo-realistic frames. Therefore, our method can control expressions by textures and decouple expressions from lip motion.
We propose an audio submodule to generate 3D face animation (represented by 3DMM parameters) from the input audio, where the 3D face is used to sample the dynamic neural textures from texture space into screen space to generate photo-realistic frames. One challenge in our method is how to decouple lip motions from expressions. Our idea is using the dynamic neural textures to represent and control expressions, and using the 3D face geometry to represent and control lip motions. To achieve this decoupling, the 3D face geometry should keep neutral\footnote{Note that neural textures only need coarse geometry.}; in other words, the 3D face does not contain expression information. Accordingly, we propose a decoupling network to transfer 3D faces with different expressions to a neutral face.

%\yzp{For achieving that,  We propose a method with 3 steps to achieve the task: (1) we use Wav2Lip~\cite{Wav2Lip}, i.e. a method for audio-driven talking-face video generation, to generate talking-face videos from the input audio and the input frames, (2) we use WM3DR~\cite{zhang2021weakly}, i.e. a method for 3D face reconstruction, to reconstruct 3D faces from the generated talking-face videos to obtain a 3D face animation, (3) we propose a decoupling network to transfer the 3D faces with expressions to neutral faces.}

We also pays attention to the texture quality in the teeth area, which is essential for high-quality talking-face videos. Our method represents face geometry by  the 3DMM model. However, 3DMM cannot provide information in the teeth part. To address this problem, we propose a teeth submodule to complete the missing texture information in the teeth area. The teeth submodule focuses on the teeth area of sampled neural textures by an affine transformation and uses a CNN to infer the missing information. Our results show that using the teeth submodule can improve the quality of the teeth area.

In summary, the main contributions of our work include: 
\begin{itemize}
\item We propose novel dynamic neural textures to generate talking-face videos with continuously controllable expressions in real time, without using training data with continuous intensity values.
\item We propose a decoupling network to transfer 3D faces with expressions to neutral faces.
\item We propose a teeth submodule to complete the missing information in the teeth area for achieving fine talking-face details and realistic textures.
\end{itemize}

\section{Related Works}

\subsection{Audio-Driven Talking-Face Video Generation}
%~\cite{ChungJZ17, SongZLWQ19, WilesKZ18, ZhouLLLW19, VougioukasPP19, Wav2Lip}
Audio-driven talking-face video generation is a typical task with multi-modal input, which uses an audio to drive a specified face (represented by either a face photo or video) and generate a new talking-face video. Deep neural network models have been developed for this task. \cite{SongZLWQ19} proposes a conditional recurrent generation network that takes an audio and an image as input. \cite{ChungJZ17} uses two CNNs to extract features from audio and photo respectively. \cite{ZhouLLLW19} uses an auto-encoder to disentangle subject-related information and speech-related information. \cite{Wav2Lip} uses a pre-trained lip-sync discriminator to correct lip-sync errors, which improves lip synchronization of generated results. Other methods use 2D facial landmarks~\cite{ChenMDX19, zhou2020makelttalk} or 3D face models~\cite{wen2020photorealistic, thies2020neural, yi2020audio, wu2021imitating, ji2021audio} to bridge the gap between audio and visual domains, which achieve better stability and inter-frame continuity. Compared with 2D facial landmarks, the 3D face model is better to handle the large changes of head pose and provide dense guidance. %\yzp{\cite{ji2021audio} can generate talking-face videos with controllable expressions by controlling the 3DMM, which uses 3D faces to represent both expressions and lip motions.}

\begin{figure*}[ht]
\centering
\includegraphics[width=0.96\textwidth]{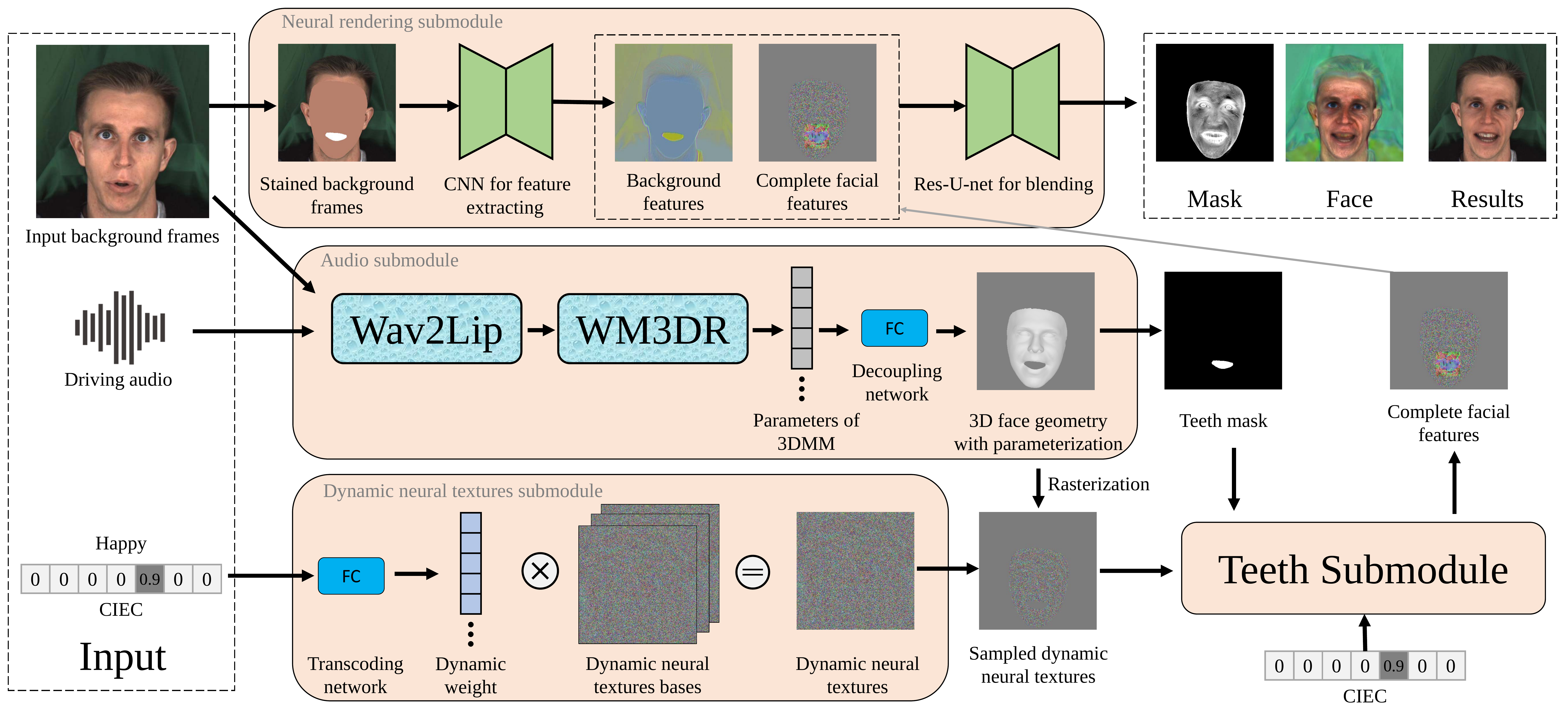}
\caption{The illustration of the proposed dynamic neural texture method. %We generate a talking-face video frame-by-frame and generate a frame from a background frame, a driving audio segment and a CIEC.
}
\label{fig:net_pipeline}
\end{figure*}

\subsection{Expression Editing}

Expression editing aims to modify the human expressions in photos or videos while keeping their identities and other characteristics unchanged, which is a special kind of image translation task. Recently, general-purpose image translation methods~\cite{choi2018stargan, richardson2021encoding} have been successfully applied in this area. However, these methods cannot completely decouple expressions from other attributes, such as lip motion and pose, which are important in talking-face videos, and thus cannot generate high-quality results. Some methods~\cite{ma2019real, geng20193d, wu2020leed, ding2018exprgan} are specially designed for expression editing. \cite{ding2018exprgan} designs an expression controller module to encode expressions as real-valued intensity vectors and other methods take only discrete intensity levels of expressions into consideration. However, \cite{ding2018exprgan} cannot be used to edit talking-face videos, because they cannot maintain lip shape synchronization with speech. In recent years, several methods~\cite{karras2017audio, cudeiro2019capture} have been proposed for generating 3D face animations with controllable expressions. Although 3D face animations can be finely controlled with high quality, they are still far from photo-realistic and are easily distinguishable from real videos.

\subsection{Neural Rendering}

Neural rendering~\cite{tewari2020state, thies2019deferred, mildenhall2020nerf} is a novel rendering technology utilizing neural networks. In contrast to traditional rendering that uses empirical models or physically-based models, neural rendering takes full advantage of neural networks and can achieve more realistic results. One important class of neural rendering technologies are neural textures~\cite{thies2019deferred}, which apply learned textures to 3D meshes to represent a scene. Neural textures have been introduced into generating high-quality talking-face videos \cite{thies2019deferred, thies2020neural}. However, existing methods only use static textures which cannot be used to model time-varying expressions. Inspired by dynamic textures~\cite{olszewski2017realistic,nagano2018pagan} which are per-frame textures, we overcome this limitation and propose dynamic neural textures, which are independent of geometry for decoupling lip motion and expressions.

\section{Our Method}

\subsection{Overview}

Our work is based on an texture-containing-expression-information observation, which is explained in Sec.~\ref{sec:where_are_exp}. The pipeline of the proposed dynamic neural texture method is illustrated in Fig.~\ref{fig:net_pipeline}, which contains 4 submodules: dynamic neural texture submodule (Sec.~\ref{sec:dynamic_neural_textures}), audio submodule (Sec.~\ref{subsec:audio_submodule}),  teeth submodule (Sec.~\ref{subsec:teeth}) and neural rendering submodule (Sec.~\ref{subsec:neural_rendering_submodule}). The input to our system is a driving audio, a sequence of expressions and a sequence of background frames (or only one background frame). Our system outputs a talking-face video with continuously controllable expressions, in which each video frame is generated from a background frame, a driving audio segment and a continuous intensity expression coding (CIEC) vector. We use two submodules --- an audio submodule and a dynamic neural texture submodule --- to decouple lip motions from expressions. The former generates 3D face animation (represented by 3DMM parameters) from input audio, and the latter obtains dynamic neural textures from CIEC, which is used to fuse features of the input expressions. In Sec.~\ref{subsec:texture_mapping}, we present the details of how to use generated 3D faces to sample dynamic neural textures. Finally, we blend the facial area into the background (Sec.~\ref{subsec:neural_rendering_submodule}), by using a CNN to extract features from background frames and using a U-Net with residual blocks from the background features and the complete facial features to simultaneously generate a color facial image and an attention mask.

\begin{figure}[t]
\centering
\includegraphics[width=\columnwidth]{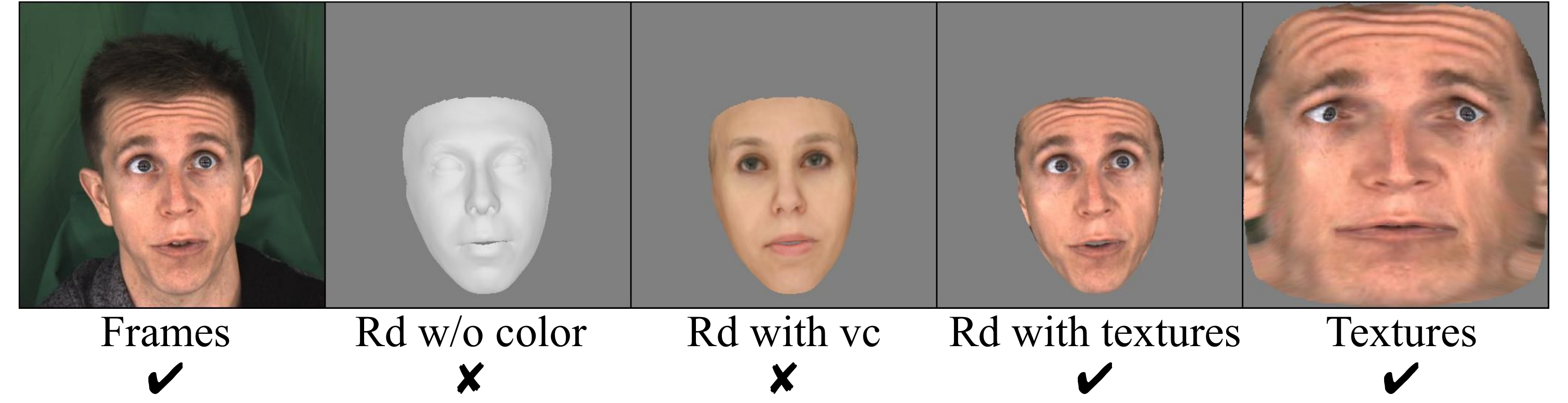}
%\vspace{-0.25in}
\caption{Illustration of five input settings, and each input setting has an indicator to show whether it contains high frequency textures. Rd means rendered and vc means vertex color. 
}
\label{fig:where_exp}
%\vspace{-0.05in}
\end{figure}

\begin{table}[t]
\centering
\caption{Cross entropy (CE) losses of five inputs. The numbers in brackets are ranks. Rd means rendered and vc means vertex color.}  
\vspace{-0.05in}
\label{tab:resnet50_expression}
%\footnotesize
\begin{tabular}{c|c|c|c}
\hline
Groups & Inputs  & CE of level $\downarrow$ & CE of type $\downarrow$ \\ \hline
\multirow{3}{*}{with textures} & Frames & 0.0338 (1) & 0.0019 (3)\\ \cline{2-4} 
& Rd with textures  & 0.0373 (2) & 0.0016 (2)\\ \cline{2-4} 
& Textures & 0.0383 (3) & 0.0012 (1) \\ \hline
\multirow{2}{*}{w/o textures} & Rd w/o color & 0.2048 (5)& 0.0326 (4) \\ \cline{2-4} 
& Rd with vc    & 0.1697 (4) & 0.0438 (5)\\ \hline
\end{tabular}
\normalsize
\vspace{-0.1in}
\end{table}

\subsection{Where Are Expressions? }
\label{sec:where_are_exp}

We use a 3DMM (with 35,709 vertices in our experiment), which has been widely used in the talking-face video generation~\cite{wen2020photorealistic, yi2020audio}, to represent the face geometry and texture in an image. The expression is a characteristic of face information and thus can be captured in geometry and texture. To explore which component captures most of the expression information, we design an experiment. We train a classification model with the same architecture (ResNet-$50$~\cite{HeZRS16}) for five different input settings (shown in Fig.~\ref{fig:where_exp}), i.e., video frames~(Frames), images rendered without color~(Rd w/o color), images rendered with vertex color~(Rd with vc), images rendered with texture map~(Rd with textures), and texture maps~(Textures). The geometry and vertex color are reconstructed from a frame by WM3DR~\cite{zhang2021weakly}, and the textures are sampled from the frame. In this experiment, we use all frontal-view videos of an actor from the MEAD dataset~\cite{wang2020mead}, which contain expressions of eight types and three intensity levels. We divide them into the training and testing sets randomly by $10$-fold cross-validation. For each input setting, two classification models are trained for expression types and intensity levels, respectively.

The five inputs (as shown Fig.~\ref{fig:where_exp}) can be divided into two groups by whether it includes high frequency textures: 1) the group with textures~(i.e., Frames, Rd with textures and Textures); 2) the group without textures~(i.e., Rd w/o color and Rd with vc). The cross entropy loss values for classifying expression types and intensity levels are shown in Table~\ref{tab:resnet50_expression}. The results show that the loss values of the group without textures are significantly higher than the other group. It indicates that given finite non-high resolution geometry~(face geometry represented by 3DMM), most expression information lies in the textures. Therefore, textures are essential for describing expressions in talking-face video generation. Neural rendering using neural textures has been proposed for generating high-quality talking-face videos from a coarse geometry representation~\cite{thies2019deferred, thies2020neural}. However, they use static textures which are not suitable for generating talking-face videos with {\it time-varying} expressions of different types and intensities. In computer graphics, dynamic textures have been proposed for dynamic 3D avatar~\cite{nagano2018pagan}, which uses different textures for different expressions. Inspired by the success of dynamic textures, we propose a novel method called {\it dynamic neural textures}, to generate photo-realistic talking-face videos with continuously controllable expressions.

\subsection{Dynamic Neural Textures}
\label{sec:dynamic_neural_textures}

Neural textures~\cite{thies2019deferred, thies2020neural} are a set of learnable feature maps in the texture space, which are used in a neural rendering step. Intuitively, the sampled neural textures could be regarded as feature maps of the screen space and the neural rendering is a neural network to transform the feature maps into a photo-realistic frame.  The set of feature maps and the neural network for neural rendering are trained in an end-to-end manner. 

Existing neural textures use static texture information during inference, which we refer to as {\it static neural textures}. Static neural textures are able to generate talking-face videos by (1) controlling the geometry of the 3D face model, whose parameters are inferred from an input audio sequence, and (2) using the 3D face model to sample the neural textures from texture space into screen space by a UV map. Different from the static neural textures which are fixed during inference, our proposed dynamic neural textures are variable and depend on the input expression. Therefore, dynamic neural textures are more expressive for performing different expression types and intensities. 

\textbf{Interpretation from the perspective of set approximation.}
Static neural textures generate talking-face videos with a fixed expression. Therefore, using different static neural textures can represent different expressions and dynamic neural textures can be regarded as an approximation for a set of static neural textures. Inspired by dynamic convolution kernels~\cite{ye2022DCKs}, we can understand dynamic neural textures in the following way. Denote by $\mathcal{E}$ the space of all expressions. For a fixed expression $t \in \mathcal{E}$, we can learn a static neural texture to represent it. For different expressions sampled in $\mathcal{E}$, we learn different static neural textures to represent them, which is a finite set of static neural textures. On the other hand, our method infers the dynamic neural textures from different expressions, and these textures play the same role of the set of static neural textures. However, the set of different static neural textures can only represent discretely sampled expressions in $\mathcal{E}$. As a comparison, dynamic neural textures can represent expressions in a continuous space and then provide a better tool to continuously control expressions.

\textbf{Continuous Intensity Expression Coding.}  For controlling expressions in talking-face videos continuously, we propose {\it continuous intensity expression coding} (CIEC) which is a continuous version of the one-hot encoding. To describe $C$ types of expressions (except the neutral expression), we define the CIEC as a $C$ dimensional vector. Similar to one-hot encoding, the CIEC allows at most one non-zero element. Each dimension characterizes an expression type and the zero vector represents the neutral expression. The value of the non-zero element represents the intensity of the corresponding expression. In our experiment, the intensity is normalized to $[0, 1]$, where $0$ indicates the neutral expression and $1$ indicates the highest intensity of that expression. In the MEAD dataset~\cite{wang2020mead}, there are three levels (level 1, 2 and 3) for each expression type. We map level 1 to $0.33$, level 2 to $0.67$ and level 3 to $1$.

We infer dynamic neural textures from the input CIEC vector (as shown in the bottom row of Fig.~\ref{fig:net_pipeline}). Since texture space is well aligned and each position in texture space has a fixed (or similar) semantic meaning, e.g., a pixel in the mouth region always corresponds to the mouth, blending a set of neural textures bases (which are learnable parameters) is a direct and effective way to obtain dynamic neural textures. We propose a transcoding network (implemented as a fully connected network) to infer a vector of weights from CIEC, and use these weights to linearly combine the neural textures bases. We implement the dynamic neural textures bases on a linear layer, which follows the transcoding network. Hence we can implement the transcoding network and the dynamic neural textures bases together as a fully connected network. Due to the continuity of the fully connected network, we obtain continuous dynamic neural textures and then continuously control the intensity levels of expressions.

Dynamic neural textures are used in the submodule of neural rendering. In particular, we use the audio submodule to generate 3D face animation (represented by 3DMM parameters) from the input audio and then use the 3D face geometry with parameterization (as UV coordinates) to sample the dynamic neural textures from texture space into screen space to generate photo-realistic frames.

\subsection{Audio Submodule}
\label{subsec:audio_submodule}

The audio submodule generates 3D face animation (represented by 3DMM parameters) from the input audio, where the 3D face is used to sample the dynamic neural textures from texture space into screen space (in which photo-realistic frames are generate). We use the following three steps to achieve the task (as shown in the middle row of Fig.~\ref{fig:net_pipeline}): (1) we use Wav2Lip~\cite{Wav2Lip} to generate talking-face videos from the input audio and the input frames, (2) we use WM3DR~\cite{zhang2021weakly} to reconstruct 3D faces from the generated talking-face videos and then obtain a 3D face animation, and (3) we 
use a decoupling network to transfer the 3D faces with expressions to a neutral face. In these steps, the challenge is how to decouple lip motions from expressions. Our solution is using the dynamic neural textures to represent and control expressions, and using the 3D face geometry to represent and control lip motions. In this way, the 3D face geometry should not provide expression information, which means the 3D face should be neutral. Accordingly, we use the decoupling network to transfer 3D faces with different expressions to a neutral face, which is used later in the dynamic neural texture submodule. 

\textbf{Decoupling Network.}
We train a decoupling network to transfer 3D faces with different expressions to neutral faces and meanwhile keep other attributes, including lip motion and identity. The task is projecting a 3D face into the subspace of neutral faces through the direction that keeps other attributes. This task is similar to expression editing of 3D faces and it only includes one target direction. The 3D faces are represented by 3DMM parameters and the parameters of our 3DMM are a 257-dimensional vector, which contains five components $\{\boldsymbol{\alpha},\boldsymbol{\beta},\boldsymbol{\delta},\boldsymbol{\gamma},\mathbf{p}\} \in \mathbb{R}^{257}$, where $\boldsymbol{\alpha} \in \mathbb{R}^{80}$ is the identity component, $\boldsymbol{\beta} \in \mathbb{R}^{64}$ is the expression component, $\delta \in \mathbb{R}^{80}$ is the texture component, $\boldsymbol{\gamma} \in \mathbb{R}^{27}$ is the illumination component, and $\mathbf{p} \in \mathbb{R}^{6}$ is the pose component. In the decoupling network, we only consider the expression component and fix the other components. The decoupling network can be regarded as a mapping function $f_d:\mathbb{R}^{64} \to \mathbb{R}^{64}$. We implement the decoupling network as a fully connected network due to its good capacity of fitting vector mappings.

\begin{figure}[htb]
\centering
\includegraphics[width=0.96\columnwidth]{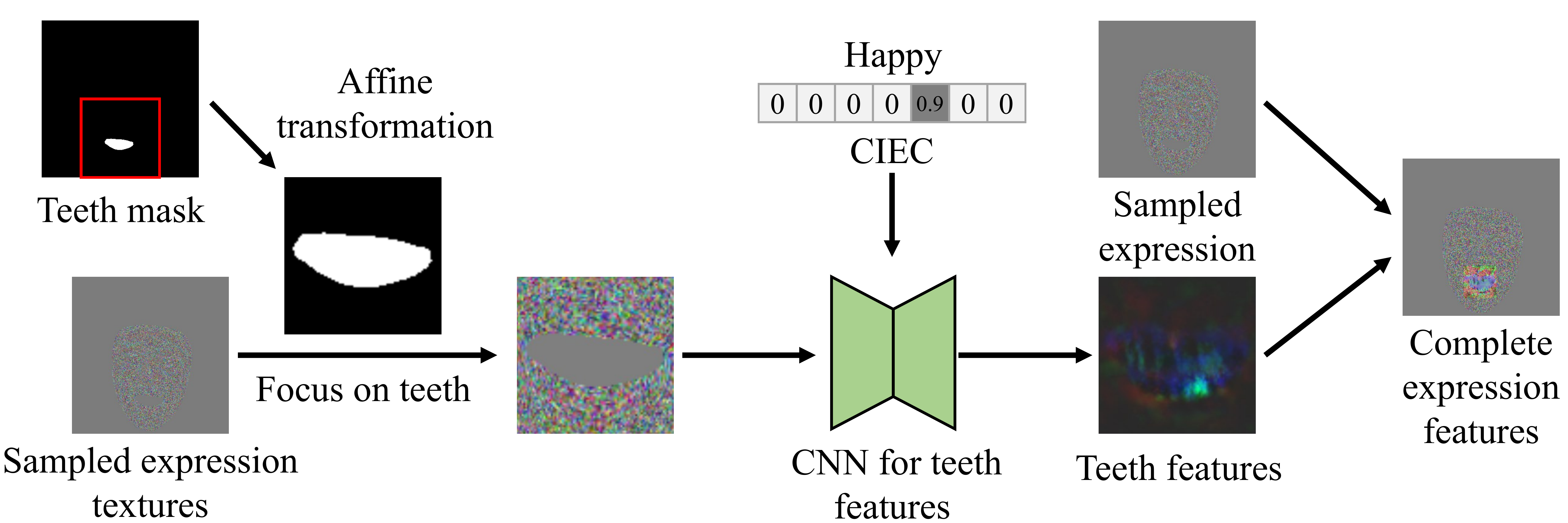}
\caption{The illustration of teeth submodule. }
\label{fig:teeth_submodule}
%\vspace{-0.1in}
\end{figure}

\subsection{Texture Mapping}
\label{subsec:texture_mapping}

In our pipeline, we use the 3D face geometry with parameterization (as UV coordinates) to sample the dynamic neural textures from texture space into screen space by rasterization. Since the faces represented by 3DMM parameters have the same topology, we have a natural parameterization of the mean face of our 3DMM and it can be used for all faces. We use the deferred rendering to implement the texture mapping: (1) we rasterize the UV coordinates to obtain a UV map in screen space, and (2) use the UV map to sample the dynamic neural textures from texture space into screen space.

\subsection{Teeth Submodule}
\label{subsec:teeth}

3DMM does not contain teeth information and then the sampled neural textures have no information about teeth. Since teeth are important in a high-quality talking-face video, we propose a teeth submodule to complete the missing texture information in the teeth area. As shown in Fig.~\ref{fig:teeth_submodule}, we use affine transformation to focus on the teeth area and use a CNN to complete teeth features. For aligning the teeth area, the affine transformation translates the center of the teeth area to the center of the image. We also use the inverse transformation to integrate the teeth features with the sampled neural textures. We apply a CNN to complete the teeth features, which is a U-Net with residual blocks. Since the teeth are related to the expression, we also input the CIEC to the CNN. We rearrange the CIEC by a fully connected network and concatenate the rearranged CIEC with the image. The output teeth features are then transformed inversely and concatenated with the sampled neural textures. 

\subsection{Neural Rendering Submodule}
\label{subsec:neural_rendering_submodule}

The neural rendering submodule in our pipeline (as shown in the top row of Fig.~\ref{fig:net_pipeline}) uses the complete facial features (including dynamic neural textures and teeth features) to generate photo-realistic talking-face video frames. Since the 3DMM does not contain hair and background, we make use of hair and background information in the input frame. We mask out the facial area and teeth area of the input frame to obtain a background frame, where the facial area and teeth area are the projection of the 3DMM reconstructed from the frame. We then use a CNN to extract background features from the masked background frames, concatenate them with the complete facial features and input them into a U-Net with residual blocks for blending. In this step, the output contains a colored facial image and an attention mask. We use the attention mask to blend the colored facial image with the background frame.

\textbf{Blending.}
We use the attention mask to blend the colored facial image with the background frame. The attention mask $\alpha_i$ is a grayscale image and the facial image $F_i$ is a color image, where $i$ indicates the $i$-th frame. Denote by $B_i$ the $i$-th background frame, the $i$-th synthetic frame (final output) $I_i'$ is calculated as:
\begin{equation}
I_i' = B_i \otimes (1 - \alpha_i) + F_i \otimes \alpha_i,
\end{equation}
where $\otimes$ is pixel-wise multiplication.

\section{Training Details}

Multiple network structures are used in our pipeline (shown in Fig.~\ref{fig:net_pipeline}). The networks of Wav2Lip~\cite{Wav2Lip} and WM3DR~\cite{zhang2021weakly} are the pre-trained models. We pre-train the decoupling network in the audio submodule as a preprocess and then all networks in the audio submodule are pre-trained models. We end-to-end train all the other networks in our pipeline, including networks of dynamic neural textures submodule, teeth submodule and neural rendering submodule.

\begin{figure*}[th]
\centering
\includegraphics[width=0.96\textwidth]{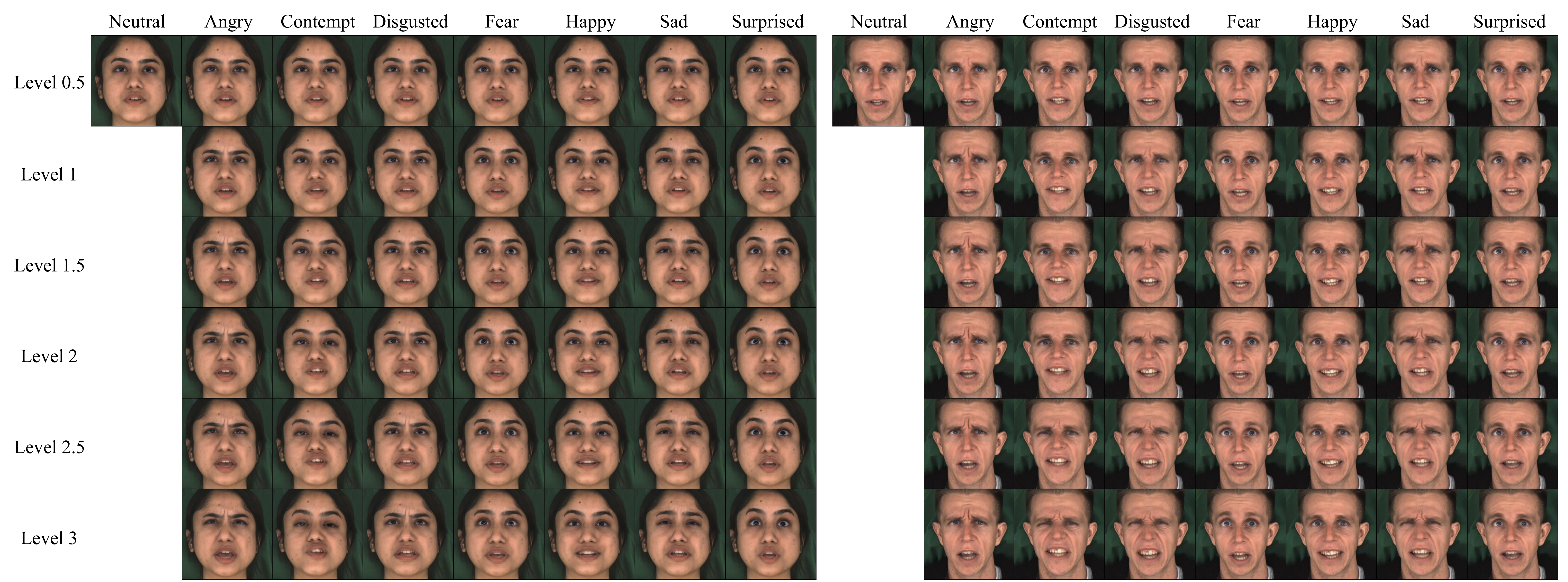}
\caption{Results with different types and sampled intensity levels of expressions. The levels 0.5, 1.5 and 2.5 do not appear in the training set. The results are %corresponding to the
consistent with the input
 expression types and intensity levels. More results are presented in the demo video.}
\label{fig:exp_level}
\end{figure*}

\subsection{Pre-training the Decoupling Network}

We present the decoupling network in Sec.~\ref{subsec:audio_submodule}. This network projects a 3D face to the subspace of neutral faces and meanwhile keeps the lip motion. It is a mapping $f_d:\mathbb{R}^{64} \to \mathbb{R}^{64}$ and implemented by a fully connected network. We reconstruct 3D faces from MEAD dataset~\cite{wang2020mead} which contains talking-face videos with different expressions (including neutral expressions), and use them to train the decoupling network. We use three loss terms to train the decoupling network, i.e. adversarial loss, neutral loss and landmarks loss.

Inspired by GAN~\cite{GoodfellowPMXWOCB14}, we use a discriminator $D$ to learn the subspace of neutral faces and ensure the output of the decoupling network is a neutral face. The adversarial loss term is:
\begin{equation}
\begin{array}{l}
L_{adv}(f_d, D) = \mathbb{E}_{F_1 \sim \mathcal{F}_{n}, F_2 \sim \mathcal{F}}(\log D(F_1) + \log(1-D(f_d(F_2)))),
\end{array}
\end{equation}
where $\mathcal{F}$ is the space of 3D faces and $\mathcal{F}_{n}$ is the subspace of neutral faces.

The decoupling network does not change the neutral face. If the input is a neutral face, the output should be the same as the input. Then we design the neutral loss term as:
\begin{equation}
\begin{array}{l}
L_{neutral}(f_d) = \mathbb{E}_{F \sim \mathcal{F}_{n}}(\lVert f_d(F) - F \rVert_1).
\end{array}
\end{equation}

Furthermore, we use mouth landmarks in input and output to constrain that the lip motions are unchanged. Accordingly, we design the landmarks loss as:
\begin{equation}
\begin{array}{l}
L_{landmarks}(f_d) = \mathbb{E}_{F \sim \mathcal{F}}(\lVert LM(f_d(F)) - LM(F) \rVert_2),
\end{array}
\end{equation}
where $LM$ is the mouth landmarks of 3D faces.

The overall loss function is in the following form:
\begin{equation}
\begin{array}{l}
L_{total}(f_d, D) \\ = L_{adv}(f_d, D) + \lambda_{1}L_{neutral}(f_d) + \lambda_{2}L_{landmarks}(f_d),
\end{array}
\end{equation}
where $\lambda_{1}$ and $\lambda_{2}$ are the weights for balancing the multiple objectives. For all experiments, we set $\lambda_{1} = 1$ and $\lambda_{2} = 50000$.

\subsection{Training Our Model End-to-end}

We train our model end-to-end in a supervised way. Our model generates talking-face videos frame-by-frame. The input to the model $g(B, F, E)$ contains a background frame $B$, a 3D face $F$ and a CIEC vector $E$, where the background frame is obtained from the ground-truth frame (i.e., the frame from the MEAD dataset, which are not generated from the audio by Wav2lip), and the 3D face is obtained from the ground-truth frame by 3D face reconstruction~\cite{zhang2021weakly} and the transferring in the decoupling network. In the training process, we skip the Wav2lip in the audio submodule and use the ground-truth frame.%, because the ground-truth frame is already the taking-face video corresponding to the audio.

Our model generates a photo-realistic frame, which should be the same as the ground-truth frame. To model this constraint, we use a perceptual loss based on a pre-trained VGG-$19$ network~\cite{simonyan2014very} $\mathcal{V}$:
\begin{equation}
\begin{array}{l}
L_{vgg}(g, I_{gt}, E_{gt}) = \lVert \mathcal{V}(g(B_{gt}, f_d(F_{gt}), E_{gt})) - \mathcal{V}(I_{gt})\rVert_1,
\end{array}
\end{equation}
where $I_{gt}$ is the ground truth frame, $F_{gt}$ is the reconstructed face from $I_{gt}$, $f_d$ is the pre-trained decoupling network, $B_{gt}$ is the background of $I_{gt}$, and $E_{gt}$ is the expression type and intensity label of $I_{gt}$.

\section{Experiments}

\subsection{Implementation Details}

We implemented our method and baseline methods with PyTorch~\cite{paszke2017automatic} and Pytorch3D~\cite{ravi2020pytorch3d}. We trained and tested the model on a server with a NVIDIA Tesla A100 GPU. We use the Adam solver with $\beta_1 = 0.9, \beta_2 = 0.999$ and learning rate of $1e^{-5}$ to optimize our model.

\textbf{Training Set.}
We train a model for a specific person by using a set of videos with different expression types and intensity levels of this person. To satisfy the need, we use MEAD dataset~\cite{wang2020mead}, which has eight expression types (neutral, angry, contempt, disgusted, fear, happy, sad, surprised) and three intensity levels (level 1, 2, 3) for each actor. To train a model for a person in MEAD, we use all frontal-view video clips of the person, about 600-700 clips and 30-40 minutes in total. The length of each clip ranges from 1 to 7 seconds. Each video clip has an expression type label and an intensity level label. 

\textbf{Running Time.}
For each batch with 8 frames, our method takes 245 ms in average, i.e., $31$ ms per frame, which is over $30$ fps. Therefore, our method can generate talking-face videos with continuously controllable expressions in real-time.

\subsection{Study on Expression Intensity Levels}

We generate results by fixing an input audio segment and using different expression types and intensity levels, as shown in Fig.~\ref{fig:exp_level}. We use all 8 expression types in the training set and 6 sampled intensity levels, i.e. 0.5, 1, 1.5, 2, 2.5, 3, where level 0.5, 1.5, 2.5 do not appear in the training set. It is observed that all the results are consistent with specified expression types and intensity levels. More results with {\it continuously varied expressions} are presented in the demo video. These results demonstrate that our method can continuously control the expressions of generated results.

\begin{figure}[htb]
\centering
\includegraphics[width=0.9\columnwidth]{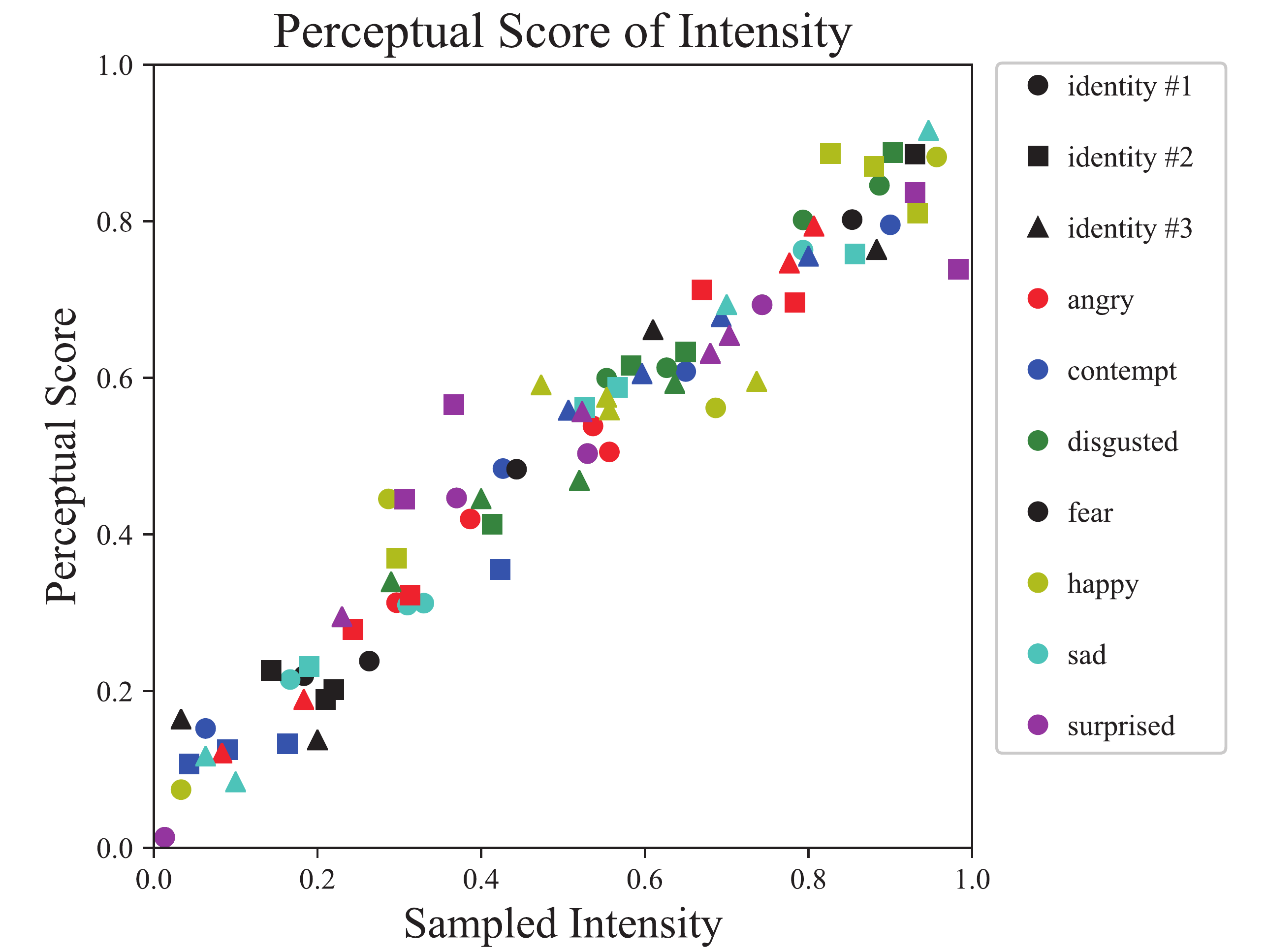}
\caption{The perceptual scores for expression intensity of randomly sampled videos, where seven colors represent seven types of expressions and three shapes represent the three identities. That the scatter is close to $y=x$ shows our method has good performance. }
\label{fig:intensity_score}
\end{figure}

\begin{figure*}[htb]
\centering
\includegraphics[width=0.96\textwidth]{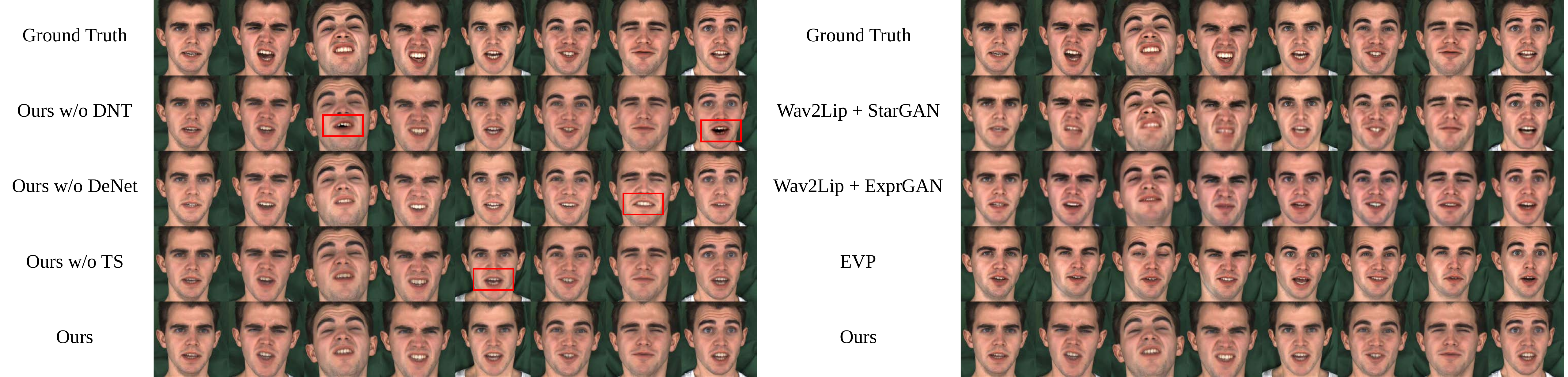}
\caption{Results and the qualitative comparison of the ablation study methods, baseline methods, EVP~\cite{ji2021audio} and our method. The left includes the ablation study methods and the right includes existing methods, where all the methods use the same set of inputs.}
\label{fig:methods_cmp}
\end{figure*}

\begin{table*}[htb]
\centering
\caption{Several metrics for expressions and lip motions of different methods, i.e. methods in ablation study, baseline methods, EVP~\cite{ji2021audio} and our method, which shows that our method achieves a good balance between various criteria. }  
\label{tab:quantitative_comparison}
\begin{tabular}{c|cccccc}
\hline
Methods & PSNR $\uparrow$ & SSIM $\uparrow$ & CE of intensity levels $\downarrow$ & CE of expression types $\downarrow$ & LMD $\downarrow$ & CSS $\uparrow$  \\ \hline
Wav2Lip~\cite{Wav2Lip} + StarGAN~\cite{choi2018stargan}  & 27.24 & 0.73 & 0.26 & 0.81 & 0.86 & 5.65\\ 
Wav2Lip~\cite{Wav2Lip} + ExprGAN~\cite{ding2018exprgan}  & 28.93 & 0.79 & 0.48 & 0.023 & 0.44 & 4.38 \\ 
EVP~\cite{ji2021audio} & 29.53 & 0.71 & 0.27 & 0.36 & 0.49 & 4.16 \\ 
Ours w/o DNT & 29.64 & 0.89 & 0.034 &  0.23 & 0.45 & {\bf 6.07}\\ 
Ours w/o DeNet & 29.99 & 0.87 & 0.054 &  {\bf 0.013} & 0.44 & 4.19\\ 
Ours w/o TS & 29.71 & 0.90 & 0.034 & 0.27 & 0.43 & 5.79\\ 
Ours & {\bf 30.39} & {\bf 0.91} & {\bf 0.029} & 0.18 & {\bf 0.39} & 5.81 \\ \hline
\end{tabular}
\end{table*}

\textbf{Perceptual Study.} 
We design a perceptual study to validate if the expressions in the generated results are consistent with the input expression labels. We ask participants to sort a set of videos with the same expression type according to intensity levels. We randomly sample $m=4$ intensities in $[0, 1]$ for each expression type and use the same audio to generate $m$ videos. We let participants drag and sort them according to intensity levels from weak to strong. We design a {\it perceptual score of intensity} based on the collected orders to measure the perceptual expression intensity of a video. Based on the collected rankings, we define a {\it perceptual intensity score}, to represent the perceptual intensity level: Denote the sampled $m$ intensities from low to high as $x = \{ x_1, \dots, x_m \}$. The order matrix $M_i$ for the order of the $i$-th participant is defined as a $m \times m$ matrix where each row and column has only one element with value $1$ and the remaining elements with value $0$. The element at $(i, j)$ with value $1$ means that the participant thought the video with $x_i$ intensity ranks $j$ in order of intensity level. The order matrix is an identity matrix if the sorted order by the participant is corresponding to the order of the sampled intensities. The average order matrix $M$ of all $k$ participants is calculated as $M = \sum_{i=1}^k M_i/k$. The perceptual intensity scores of $x$ is $s = Mx$. The score will be equal to the sampled intensity if $M$ is an identity matrix, i.e. the sorted order by all participants are corresponding to the order of the sampled intensities. $16$ participants are recruited to attend the perceptual study and each of them sorts $21$ groups of videos of $3$ identities and $7$ types of expressions, each group contains $m=4$ generated videos with $m$ different input intensity labels. The perceptual intensity scores of videos are shown in Fig.~\ref{fig:intensity_score}, where the scores are close to the input CIEC, demonstrating that the expression intensities in the generated results are consistent with the input expression intensity labels.

\subsection{Metrics for Evaluation}

We use the following metrics to evaluate the generation results of our methods and baseline methods. The video clips in MEAD dataset~\cite{wang2020mead} are used to train and test the methods. We use the early $80\%$ duration of each video clip as the training set and the late $20\%$ duration of each video clip as the testing set to calculate the metrics.

\textbf{Metrics for Video Quality.} We use Peak Signal to Noise Ratio (PSNR) and Structural Similarity Index Metrics (SSIM) to evaluate the quality of the generated videos.

\textbf{Metrics for Expressions.} We use two classification losses for expressions (i.e. loss of intensity level and loss of expression type) to evaluate the expressions of the generated videos. The two losses are cross entropy obtained by the ResNet-50 (as shown in Sec.~\ref{sec:where_are_exp}) for classifying intensity levels and types of expressions.

\textbf{Metrics for Lip Synchronization.} We use landmarks distance (LMD)~\cite{ChenLMDX18} and confidence score of synchronization (CSS)~\cite{chung2016out} to measure the lip synchronization of the generated videos, where the averaged CSS of the ground truth in the test set is 6.71.

\subsection{Ablation Study}

To validate the effectiveness of the dynamic neural textures, the decoupling network and the teeth submodule, we compare the results generated by our method with and without dynamic neural textures (DNT), decoupling network (DN) or teeth submodule (TS). The method without dynamic neural textures (ours w/o DNT) uses static neural textures (instead of DNT) and uses 3D face geometry to control the expressions, which skip the decoupling network because the decoupling network and the dynamic neural textures should work together. The method without decoupling network (ours w/o DeNet) skip the decoupling network, which uses both geometry and textures to control the expressions. The method without teeth submodule (ours w/o TS) directly skip the teeth submodule and input sampled neural textures to the neural rendering submodule. We compare our full method with them. The qualitative comparison is shown in Fig.~\ref{fig:methods_cmp} and the supplementary demo video. The quantitative comparison is summarized in Table~\ref{tab:quantitative_comparison}. The qualitative and quantitative comparisons show the dynamic neural textures and the decoupling network (i.e. using 3D faces and textures to represent lip motions and expressions respectively) are essential for generating talking-face videos with continuously controllable expressions. Moreover, the teeth submodule is helpful for generating talking-face videos with high-quality mouth region.

\subsection{Comparison with Existing Methods}

\textbf{Baseline Methods.} We propose two baseline methods by combining two steps. The first step generates talking-face videos with a neutral expression and the second step transfers the expression of each frame from neutral to the target expression. We select Wav2Lip~\cite{Wav2Lip} for the first step, which is a state-of-the-art method for generating audio-driven talking-face video. We select both StarGAN~\cite{choi2018stargan} and ExprGAN~\cite{ding2018exprgan} as the second step, where StarGAN is a state-of-the-art method for multi-domain image translation and ExprGAN is specially designed for expression editing. Note that StarGAN uses discrete intensity levels and ExprGAN uses continuous intensity levels.

EVP~\cite{ji2021audio} is a state of the art for emotional talking face generation. We also compare it with our method.

The quantitative comparison are summarized in Table~\ref{tab:quantitative_comparison}. The qualitative comparison is shown in Fig.~\ref{fig:methods_cmp} and the demo video in the supplementary material. The results in the demo video show that baseline methods generate low quality results and cannot completely decouple expressions from lip motions and head poses. The results of EVP~\cite{ji2021audio} show it generates results with inaccurate lip synchronization and it cannot decouple expressions from lip motions due to it uses 3D faces to represent both lip motions and expressions. By using 3D faces and DNT, our method can effectively decouple expressions from lip motions, and well control the head pose through a 3D face. 

\section{Conclusion}

In this paper, we propose dynamic neural textures to generate talking-face videos with continuously controllable expressions in real-time. For decoupling lip motions from expressions, we propose a decoupling network. We also propose a teeth submodule to complete the missing information in the teeth area for a 3D face model with a hole in the mouth. Quantitative and qualitative results show our method can generate high quality results with continuously controllable expressions. 

% \begin{acks}
% \end{acks}

\bibliographystyle{ACM-Reference-Format}
\bibliography{main}

\end{document}